\documentclass[conference]{IEEEtran}
\usepackage{times}

\usepackage[numbers]{natbib}
\usepackage[colorlinks=true]{hyperref}
\usepackage{amsmath}
\usepackage{graphicx}
\usepackage{subcaption}
\usepackage{color}
\usepackage[normalem]{ulem}

\hypersetup{pdftitle={PVEs: Position-Velocity Encoders for Unsupervised Learning of Structured State Representations},
pdfauthor={Rico Jonschkowski and Roland Hafner and Jonathan Scholz and Martin Riedmiller}}

\newcommand{\E}{{\rm I\kern-.3em E}}
\usepackage{balance}
\IEEEoverridecommandlockouts

\begin{document}

\title{{\bf PVE}s: {\bf P}osition-{\bf V}elocity {\bf E}ncoders for Unsupervised Learning of Structured State Representations}
\author{\authorblockN{Rico Jonschkowski\textsuperscript{1,2}, Roland Hafner\textsuperscript{1}, Jonathan Scholz\textsuperscript{1}, and Martin Riedmiller\textsuperscript{1}}\authorblockA{\textsuperscript{1}DeepMind, \textsuperscript{2}Robotics and Biology Laboratory at Technische Universit\"at Berlin}}



\maketitle


\begin{abstract}
We propose position-velocity encoders (PVEs) which learn---without supervision---to encode images to positions and velocities of task-relevant objects. PVEs encode a single image into a low-dimensional position state and compute the velocity state from finite differences in position. In contrast to autoencoders, position-velocity encoders are not trained by image reconstruction, but by making the position-velocity representation consistent with priors about interacting with the physical world. We applied PVEs to several simulated control tasks from pixels and achieved promising preliminary results.
\end{abstract}

\section{Introduction}


While position and velocity are fundamental components of state representations in robotics, robots cannot directly sense these properties. Instead, they need to extract task-relevant position and velocity information from sensory input. For robots to be versatile, they must be able to learn such representations from experience. Deep learning allows to learn position-velocity representations in principle---but most existing approaches depend crucially on labelled data.

In this paper, we investigate how robots could learn position-velocity representations without supervision. We approach this problem by using robotics-specific prior knowledge about interacting with the physical world, also known as \emph{robotic priors} \citep{Jonschkowski-14-AR}, in order to learn an encoding from high-dimensional sensory observations to a low dimensional state representation. Our contribution is to split the state representation into a velocity state and a position state and to incorporate robotic priors about position and velocity in the form of model constraints and learning objectives. 


Our method, the \emph{position-velocity encoder}~(PVE), implements a hard model constraint by estimating velocity states from finite differences in position states. This constraint fixes the relation between these two parts of the state representation. Additionally, PVEs include soft objectives that measure consistency with robotic priors. These objectives are optimized during learning and shape which information is encoded and how multiple state samples relate to each other. Both ingredients work together to learn an encoding into a structured state representation that includes \emph{position states}, which describe information from a single observation, and \emph{velocity states}, which describe how this information changes over time.

\begin{figure}[t]
    \centering
    \includegraphics[width=\columnwidth]{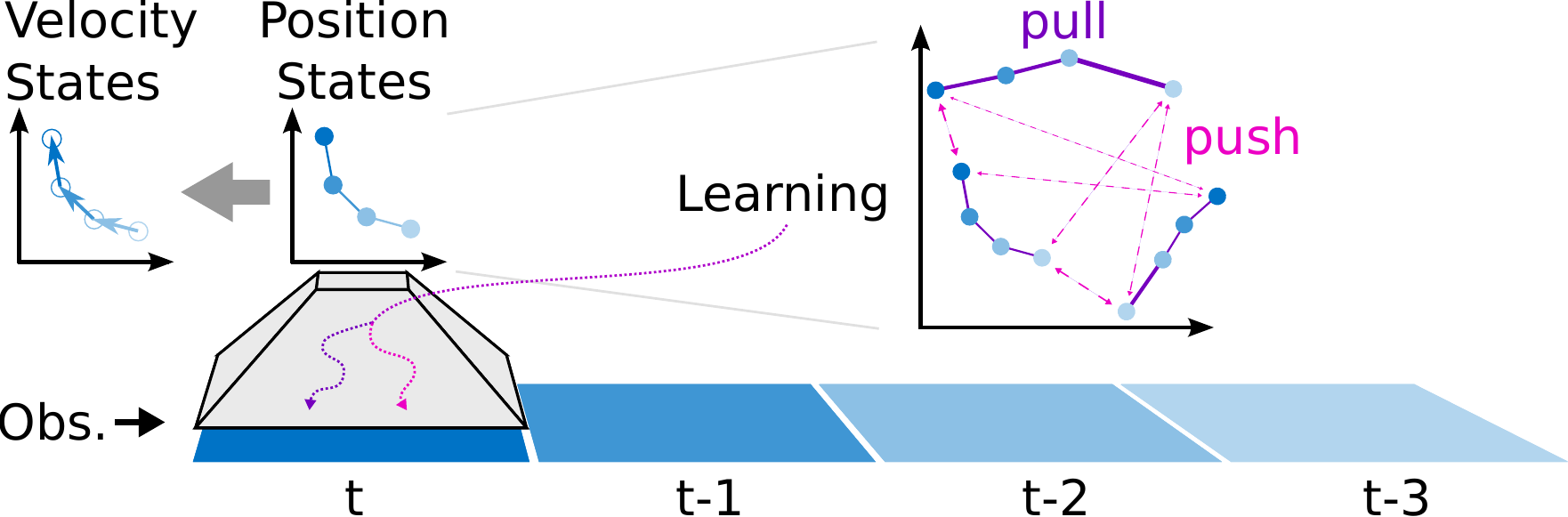}
    \caption{PVEs encode an observation into a low-dimensional position state. From a sequence of such position states, they estimate velocities. PVEs learn the encoding by optimizing consistency of positions and velocities with robotic priors.}
    \label{fig:overview}
\end{figure}

Figure~\ref{fig:overview} shows the position encoder that maps observations (blue rectangles) to position states (blue dots). The velocity state---the time derivative of the position state---is approximated from finite differences in position. This structured state space allows us to formulate new robotic priors, specifically for positions and velocities, in the form of learning objectives.

PVEs learn to encode observations into states without state labels and without learning a decoder. Instead, they learn the encoding by making position states and their derivatives consistent with different robotic priors. Inconsistency with each prior is measured in a loss function. PVEs learn by minimizing a weighted sum of these losses using gradient descent. The gradients can be imagined as forces in the state space that pull state samples together (when they should be similar) or push them apart (when they should be different). Backpropagation transforms these forces into parameter changes in the encoder (see pink and purple arrows in Fig.~\ref{fig:overview}).

In our preliminary experiments, we apply position-velocity encoders to simulated control tasks from pixels. We show that PVEs are able to discover the topology and the dimensionality of the task, that they can learn equivalent representations from different camera perspectives, that they capture information about the true positions and velocities of physical objects in the scene, and that reinforcement learning based on the learned position-velocity state can produce precise control.

\section{Related Work}

This paper extends work on learning state representations with robotic priors by Jonschkowski and Brock~\citep{Jonschkowski-14-AR}, which introduced the idea of robotic priors and their implementation in the form of objectives for representation learning. Our extension to position-velocity states is inspired by work on physics-based priors in model-based reinforcement learning by Scholz et al.~\citep{Scholz-14}, which proposed to learn a physically plausible dynamics model given a position-velocity representation. Here, we turn their approach around and ask: How could we learn a position-velocity representation from sensory input without specifying which positions and velocities are relevant for the task?

The answer we are proposing, the position-velocity encoder, works by incorporating prior knowledge in two different ways, which fit Mitchell's categorization of \emph{inductive biases} into restriction biases and preference biases \citep[p. 64]{Mitchell-97}. \emph{Restriction biases} restrict the hypothesis space that is considered during learning, such as our constraint in the PVE model to estimate velocities from finite differences in position (rather than learning to extract velocity information from a sequence of observations). \emph{Preference biases} express preferences for certain hypothesis, such as our loss functions for training PVEs, which measure inconsistency with robotic priors.

Other examples of restriction biases in the visual representation learning literature include convolutional networks \citep{lecun1995convolutional}, spatial transformer networks \citep{jaderberg2015spatial}, and spatial softmax~\citep{LevineFDA15}, which incorporate priors about visual input as architectural constraints of a neural network, as well as backprop Kalman Filters \citep{Haarnoja-16} and end-to-end learnable histogram filters \citep{Jonschkowski-16-NIPS-WS}, which incorporate the structure of the Bayes' filter algorithm for recursive state estimation. SE3-nets~\citep{byravan2016se3} implement assumptions about rigid body transformations. While these approaches can regularize learning, unsupervised learning (also) requires suitable preference biases.

Preference biases for internal representations are commonly expressed \emph{indirectly} via other learnable functions based on the representation. For example, others train representations by using them to learn image reconstruction \citep{Watter-15}, prediction of future states \citep{Sun-16, Watter-15}, or other auxiliary tasks \citep{Jaderberg-16, Mirowski-16}. A powerful but underexplored alternative is to express preference biases \emph{directly} on the learned representation, an approach which can enable to learn symbol grounding \citep{Jetchev-13} or perform label-free supervised learning \citep{Stewart-16}. Direct preference biases are also the focus of metric learning, e.g. learning representations of faces using the fairly general triplet loss \citep{Schroff-15}, but there is large potential for formulating more informative robotic priors in the form of direct preference biases, as we do in this work.

\section{{\bf P}osition-{\bf V}elocity {\bf E}ncoders}

Position-Velocity Encoders (PVEs) learn to map raw observations into a structured state space that consists of a position part and a velocity part. PVEs are trained without state labels and they do not need to learn auxiliary functions such as reconstructing observations or predicting state transitions. PVEs achieve this by combining two key ideas:
\begin{enumerate}
    \item PVEs encode the current observation into a \emph{position state} and estimate a \emph{velocity state} from finite differences in position (more details in Sec.~\ref{subsec:model}).
    \item PVEs are trained by optimizing consistency with robotic priors about positions, velocities, and accelerations (more details in Secs.~\ref{subsec:priors}~\&~\ref{subsec:training}).
\end{enumerate}

\subsection{Model}
\label{subsec:model}

The PVE model consists of a convolutional network and a numerical velocity estimation. The convolutional network $\phi$ encodes a visual observation ${\bf o}_t$ into a low-dimensional position-state ${\bf s}^{(p)}_t$, where superscript $(p)$ stands for \emph{position}.
\begin{align*}
{\bf s}^{(p)}_t &= \phi({\bf o}_t).
\end{align*}
From the difference of the last two position states ${\bf s}^{(p)}_t$ and ${\bf s}^{(p)}_{t-1}$, the model estimates the velocity state ${\bf s}^{(v)}_t$:
\begin{align*}
{\bf s}^{(v)}_t &= \alpha({\bf s}^{(p)}_{t} - {\bf s}^{(p)}_{t-1}),
\end{align*}
where $\alpha$ is a  hyperparameter that subsumes $\frac{1}{\text{timestep}}$ and scales velocity states. It is important that velocity states have the right scale relative to position states in order to create a sensible metric in the combined state ${\bf s}_t$, which we construct by stacking the position state and the velocity state.
\begin{align*}
{\bf s}_t &= \begin{bmatrix} {\bf s}_t^{(p)} \\ {\bf s}_t^{(v)} \end{bmatrix}.
\end{align*}
We can also use finite differences to estimate acceleration (or jerk, jounce, etc.). We do not include these derivatives in the state because we assume that the robot controls accelerations by its actions. But we do use the acceleration state in some loss functions. We compute the acceleration state ${\bf s}^{(a)}_t$ in the same way as the velocity state but we omit the scaling since we will not use accelerations in the combined state space:
\begin{align*}
{\bf s}^{(a)}_t &= {\bf s}^{(v)}_{t} - {\bf s}^{(v)}_{t-1}.
\end{align*}

\subsection{Robotic Priors and Learning Objectives}
\label{subsec:priors}

The encoder $\phi$ is trained by making the combined state space consistent with a set of robotic priors, which we will describe in this section. These priors use the structured state space and are specific to positions, velocities, and accelerations. Consistency with these priors is defined in the form of loss functions that are minimized during learning. The following list of robotic priors should be understood as an exploration into this matter, not as a final answer.

\subsubsection{\textbf{Variation}} \emph{Positions of relevant things vary.} As the robot explores its task and manipulates its environment, the positions of task-relevant objects (including itself) will vary---otherwise there is not much that the robot could learn. If we assume that positions of relevant objects vary in the robot's experience, the internal representation of such positions must also vary; random pairs of position states should not be similar. Therefore, we optimize consistency with the variation prior by minimizing the expected similarity between random pairs of position states,
\begin{align*}
L_{\textrm{variation}} = \E\Big[e^{-\lVert{\bf s}_{a}^{(p)} - {\bf s}_{b}^{(p)}\rVert}\Big],
\end{align*}
where we use $e^{-\text{distance}}$ as a similarity measure that is $1$ if the distance is $0$ and that goes to $0$ with increasing distance between the position states, which is exactly what we want.

\subsubsection{\textbf{Slowness}} \emph{Positions change slowly}~\citep{SFA}. Physical objects do not teleport; they do not change their position arbitrarily from one second to the next. To make the internal position state consistent with the slowness prior, we minimize the expected squared distance between consecutive position states,
\begin{align*}
L_{\textrm{slowness}} = \E\Big[\lVert{\bf s}_t^{(p)} - {\bf s}_{t-1}^{(p)}\rVert^2\Big].
\end{align*}
Since this change in position is directly connected to the rate of position change (or velocity), we can also write down the same loss using the velocity state.
\begin{align*}
L_{\textrm{slowness}} = \E\bigg[\Big\lVert\frac{{\bf s}_t^{(v)}}{\alpha}\Big\rVert^2\bigg], 
\end{align*}
where $\alpha$ is the scaling hyperparameter defined earlier. This reformulation hints at a different interpretation of slowness, which is simply: \emph{velocities are low}\footnote{Note that defining the slowness prior to mean \emph{velocities are low} translates to the loss function $L_{\textrm{slowness}} = \E[({\bf s}_t^{(v)})^2] = \E[(\alpha({\bf s}^{(p)}_{t} - {\bf s}^{(p)}_{t-1}))^2]$, which depends on the scaling parameter $\alpha$. We use the other formulation to make this loss independent of $\alpha$ because we want to change $\alpha$ during training without affecting this loss (see Sec~\ref{subsec:training} for more details).}.

\subsubsection{\textbf{Inertia}} \emph{Velocities change slowly.} Since physical objects have inertia, they resist changes to their velocity (both in direction or magnitude). If we assume limited forces to overcome this resistance, velocities should only change by small amounts. Note how the inertia prior corresponds to the slowness prior applied to velocities. 
\begin{align*}
L_{\textrm{inertia}} = \E\Big[\lVert{\bf s}_t^{(v)} - {\bf s}_{t-1}^{(v)}\rVert^2\Big] = \E\Big[\lVert{\bf s}_t^{(a)}\rVert^2\Big].
\end{align*}
This formulation of the inertia prior focuses on large velocity changes due to the square in the loss function. Alternatively, we can define the loss function based on \emph{absolute} changes.
\begin{align*}
L_{\textrm{inertia (abs)}} = \E\Big[\lVert{\bf s}_t^{(a)}\rVert\Big].
\end{align*}
Small changes in velocity have a higher weight in the second loss compared to the first loss. We found that combining both losses leads to better results than using either one of them.

\subsubsection{\textbf{Conservation}} \emph{Velocity magnitudes change slowly.} This prior derives from the \emph{law of conservation of energy}, which states that the total energy in a closed system remains constant. As the robot applies forces to the environment, we do not have a closed system. Additionally, we cannot estimate, e.g. kinetic energy without knowing the masses of objects, let alone potential energy stored in springs etc. Still, we want to enforce the same idea of keeping the absolute amount of energy, or in our case "movement" similar in consecutive time steps.
\begin{align*}
L_{\textrm{conservation}} = \E\Big[\big(\lVert{\bf s}_t^{(v)}\rVert - \lVert{\bf s}_{t-1}^{(v)}\rVert\big)^2\Big].
\end{align*}

\subsubsection{\textbf{Controlability}} \emph{Controllable things are relevant.} The objects that can be controlled by the robot are likely relevant for its task. If the robot acts by applying forces, controllable things could be those whose accelerations correlate with the actions of the robot. Accordingly, we can define a loss function per action dimension $i$ to optimize covariance between action dimension $i$ and accelerations in a state dimension $i$.
\begin{align*}
L_{\textrm{controlability (i)}} &= e^{-\textrm{Cov}({\bf a}_{t,i}, {\bf s}_{t+1,i}^{(a)})} \\
&= e^{-\E\left[\big(a_{t,i} - \E[a_{t,i}]\big)\big(s_{t+1,i}^{(a)} - \E[s_{t+1,i}^{(a)}]\big)\right]}.
\end{align*}
Note that we used this loss in only one of the tasks---ball in cup---because the above priors were insufficient. The results for this task are still preliminary. A complete solution of this task and a deeper investigation into other formulations of controlability are part of future work.



\subsection{Training Procedure}
\label{subsec:training}
We train PVEs by minimizing a weighted sum of the loss functions described above using gradient descent. This section explains the training procedure in detail.

\subsubsection{Data Gathering} 
First, the robot gathers data by exploring its environment. Since we are using a reinforcement learning setting, the data consist of sequences of observations, actions, and rewards. Most of the presented loss functions only use observations, the controlability loss also uses actions, but none of our current losses uses the reward signal.

\subsubsection{Loss Computation}
We iterate through the collected data in mini batches, which consist of a small set of short sequences. For each mini-batch, we compute the loss functions by replacing expectations with statistical averages.\footnote{For the variation loss, we sample all pairs of experiences with the same time step in different sequences of the mini batch. For all other losses we consider all samples in the mini batch.}

\subsubsection{Loss Combination}
We combine these losses in a weighted sum. Finding the right weights is important because they balance how much each prior is enforced during learning. We determined these weights empirically by adjusting them until the gradients in the encoder parameters had similar magnitudes for all priors. Future work should try to automate this process of weight tuning, potentially by applying the same heuristic in an automated fashion.

\subsubsection{Parameter Updates}
For each mini-batch, we compute the gradient\footnote{Some of the gradients can only be computed after adding small Gaussian noise to the encoded states.} of the combined loss with respect to the encoder parameters using symbolic auto-differentiation \citep{tensorflow-15} and perform an update using the Adam optimizer \cite{Kingma-14}. We iterate this process until convergence.


\subsubsection{Velocity Scaling Curriculum}
While training PVEs, we follow a curriculum that in the beginning focuses on positions and only later also takes velocities into account. This curriculum is implemented by changing the velocity scaling parameter $\alpha$. In the first phase, we train with $\alpha=0$ until convergence. In the second phase, we increase $\alpha$ linearly from $0$ to its final value and train until convergence again. In phase one, only the first two priors, variation and slowness, are active. Surprisingly, these two are powerful antagonists that can unfold the topology of the position-state space. The second phase mainly smooths the state space such that velocities can be accurately estimated from finite differences.

\subsubsection{Hyperparameters}
We used the following hyperparemeters in our experiments. The convolutional network had three convolutional layers with 16, 32, and 64 channels, kernel size 5x5, and stride 2, followed by three fully connected layers of sizes 128, 128, and 5 (for a 5-dimensional position state). Every layer except the last one was followed by a ReLu nonlinearity \cite{Nair-10}. The mini-batch size was 32 sequences of 10 steps each. The maximum velocity scaling $\alpha$ was $10$. The weights for the different losses are shown in Table~\ref{tab:weights}.

\begin{table}[]
    \centering
    \caption{Loss weights per task (same for a and b).}
    \label{tab:weights}
    \begin{tabular}{ | l | l | l | l |}
    \cline{2-4}
    \multicolumn{1}{c|}{} 
     & Task a & Task b & Task c \\
    \hline
    Variation & 1.0 & 1.0 & 1.0 \\
    Slowness & 1.0 & 1.0 & 1.0 \\
    Inertia & 0.1 & 0.1 & 0.001 \\
    Inertia (abs) & 0.1 & 0.1 & 0.02 \\
    Conservation & 0.2 & 0.2 & 0.005 \\
    Controlability (i=1,2) & 0.0 & 0.0 & 0.5\\
    \hline
    \end{tabular}
\end{table}







\begin{figure}[h]
    \centering
    \begin{subfigure}[t]{0.32\columnwidth}
        \includegraphics[height=2.0cm]{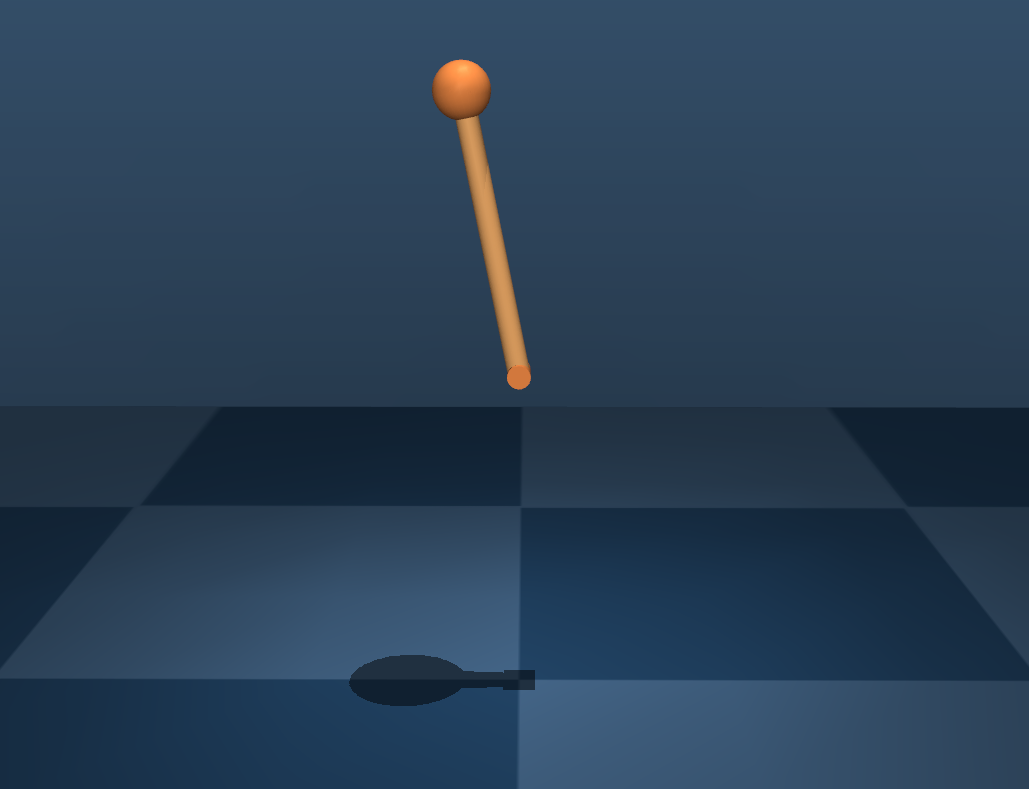}
        \caption{Inverted pendulum}
    \end{subfigure}
    \hspace{0.008cm}
    \hfill
    \begin{subfigure}[t]{0.32\columnwidth}
        \includegraphics[height=2.0cm]{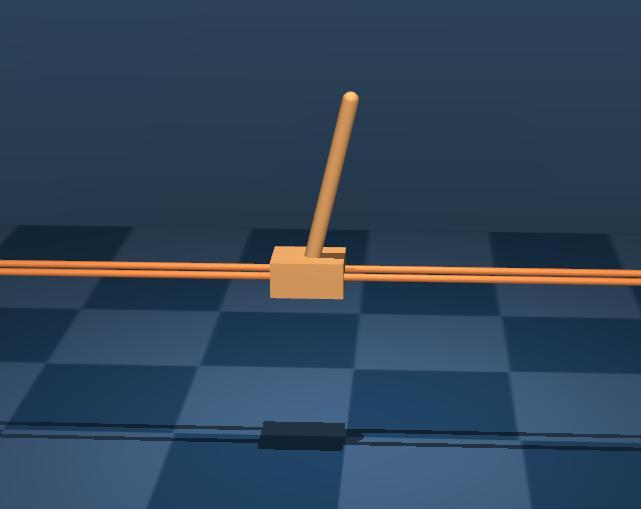}
        \caption{Cart-pole}
    \end{subfigure}
    \hfill
    \begin{subfigure}[t]{0.32\columnwidth}
        \includegraphics[height=2.0cm]{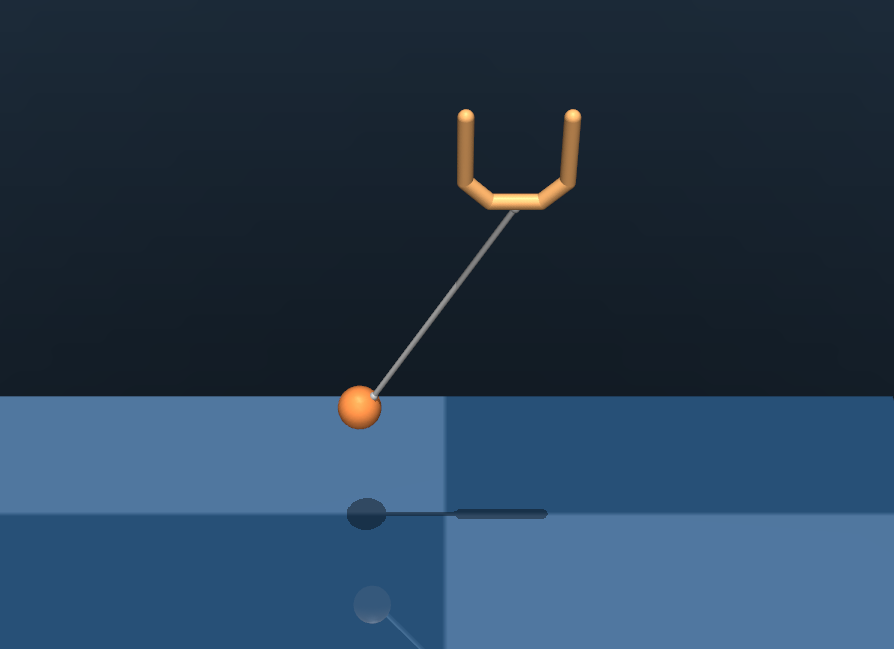}
        \caption{Ball in cup}
    \end{subfigure}
    \caption{Three control tasks from pixel input}
    \label{fig:tasks}
\end{figure}

\section{Experiments and Results}

We applied PVEs to a series of simulated control tasks from raw pixel input (see Fig.~\ref{fig:tasks}). All tasks use the MuJoCo simulator~\citep{Todorov-12}. For each task, we collected a batch of training data that consists of 1000 short trajectories of 20 steps by randomly sampling start configurations with different positions and velocities and applying a random policy.

\subsection{Tasks}

\subsubsection{Inverted Pendulum}
The inverted pendulum is a stick that is fixed to a rotary joint at one end. The goal is to swing it up and balance it upright by applying forces at the joint. However, the motor is not strong enough to pull the pendulum up all at once. Instead the pendulum must be swung back and forth to generate enough (but not too much) momentum.

\subsubsection{Cart-Pole}
The cart-pole task is an extension of the inverted pendulum task. Since the pole is attached to a cart with a passive joint, it can only be swung up by accelerating the car correctly, which requires precise control.

\subsubsection{Ball in Cup}
This task includes a cup and a ball attached to the bottom of the cup with a string. The goal is to move the cup in such a way that the ball lands in the cup. In our version of the task, cup and ball can only move in the plane.

\subsection{Learned Position-Velocity Representations}
For each task, we will now look at the learned state representations. We visualize 5-dimensional position-states by projecting to their principal components.

\subsubsection{Inverted Pendulum}
\begin{figure}[t]
    \centering
    \begin{subfigure}[t]{0.43\columnwidth}
        \includegraphics[width=\columnwidth]{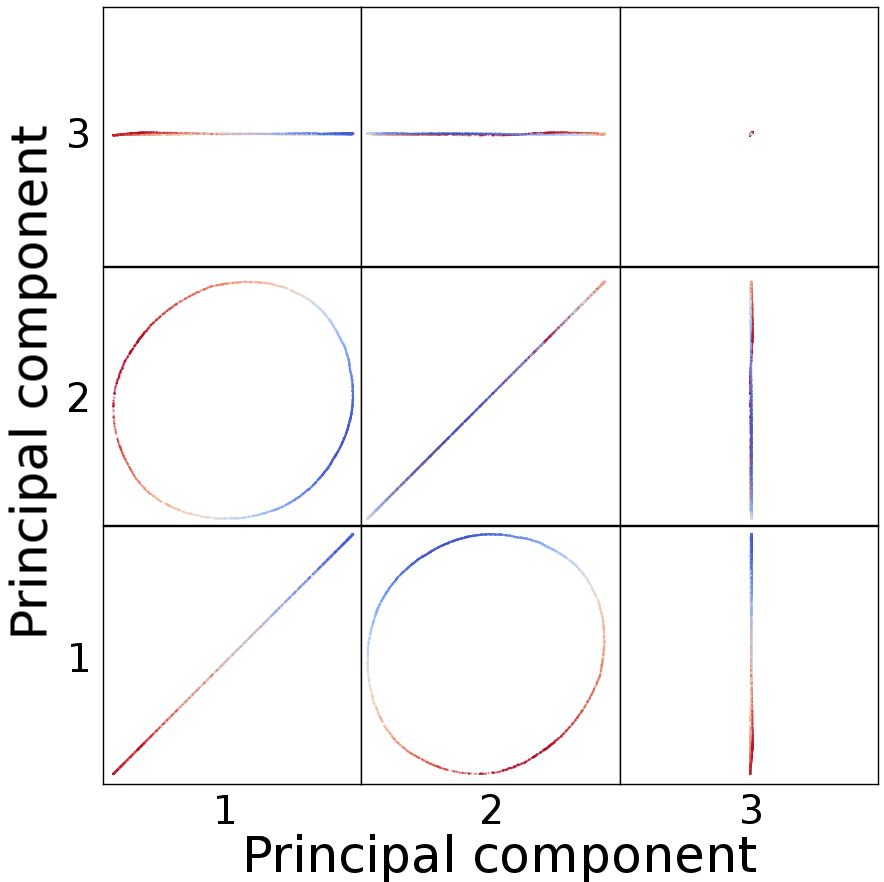}
        \caption{Encoded position states}
        \label{fig:rep_pendulum}
    \end{subfigure}
    \hfill
    \begin{subfigure}[t]{0.53\columnwidth}
        \includegraphics[width=\columnwidth]{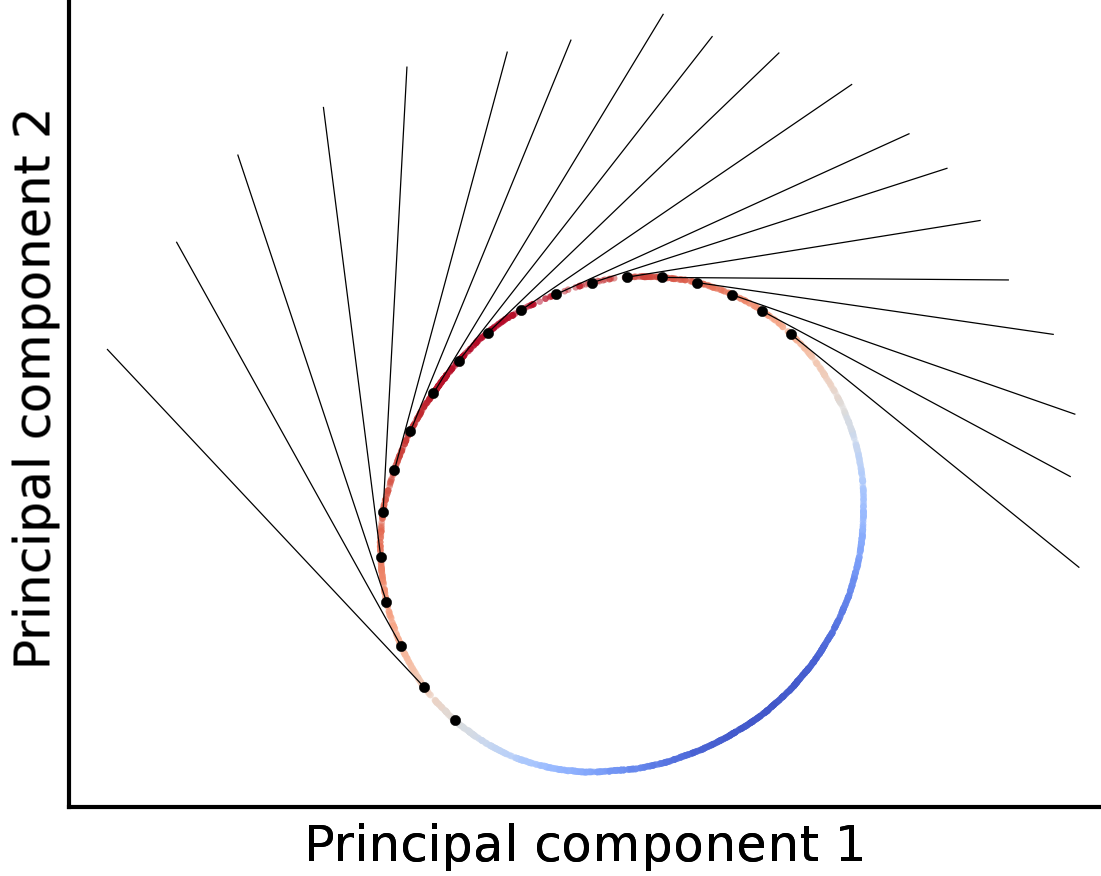}
        \caption{Overlayed state sequence}
        \label{fig:rep_seq_pendulum1}
    \end{subfigure}
    \\
    \begin{subfigure}[t]{\columnwidth}
        \vspace{0.3cm}
        \includegraphics[width=\columnwidth]{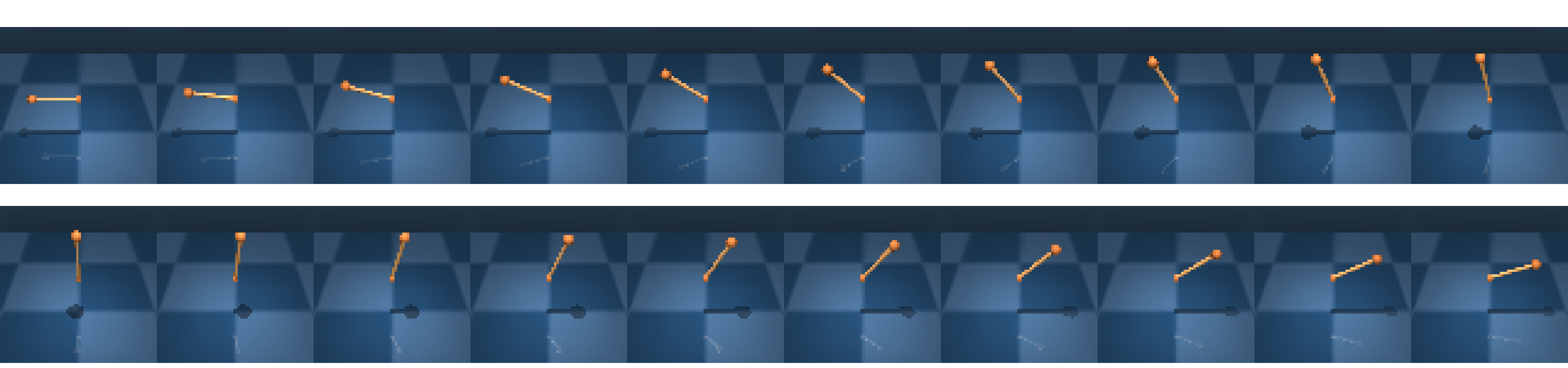}
        \caption{Observation sequence}
        \label{fig:rep_seq_pendulum2}
    \end{subfigure}
    \caption{For the inverted pendulum, PVEs learn a circular position representation that allows accurate velocity estimation. Each dot in (a) and (b) is the encoding of a single observation. The color denotes the reward received with the observation (red = high, blue = low). Black dots in (b) show the encoding of the observation sequence in (c). Black lines show the estimated velocities. {\bf Supplementary videos}: \url{http://youtu.be/ipGe7Lph0Lw} shows the learning process, \url{http://youtu.be/u0bQwz89h1I} demonstrates the learned PVE.}
    \label{fig:rep_pendulum_all}
\end{figure}

The state representation learned by the PVE is shown in Figure~\ref{fig:rep_pendulum}, where we can see the encoding of test observations into the position-state space. Each dot is the position encoding of a single image. The color denotes the amount of reward that was achieved in that instance.

The plot shows a number of interesting results. First, observations that correspond to similar rewards are encoded close together in position space. Second, the position states form a circle, which makes sense because the inverted pendulum moves in a circle. Third, all principal components after the first two are close to zero. This means that the circular encoding lies on a plane in the five-dimensional space---the PVE discovered that the task is two dimensional\footnote{Even though the task only has one positional degree of freedom (the angle of the pendulum), we need at least two dimensions if we want a Euclidean metric to make sense in this space, such that there are no jumps as from 360 to 0 degrees in an angular representation.}.

Next, we will look at the estimated velocities in the learned space. In Figure~\ref{fig:rep_seq_pendulum1}, we overlayed encoded training data colored by reward with the encoding of a single sequence of observations shown in Figure~\ref{fig:rep_seq_pendulum2}. The position states are marked with black dots and the velocity state vectors are drawn as lines. In the observation sequence, the pendulum swings from the left side to the top and then to the right side. Similarly, the encoded positions move from a medium reward region via the high-reward region (red color) to the medium reward region on the other side. During this motion, the velocity estimations are tangential to the circle in the position space with minimal noise, which should be useful for controlling the pendulum (see video links in Fig.~\ref{fig:rep_pendulum_all}). 

\subsubsection{Cart-Pole}

\begin{figure}[t]
    \centering
    \begin{subfigure}[t]{0.49\columnwidth}
        \includegraphics[width=\columnwidth]{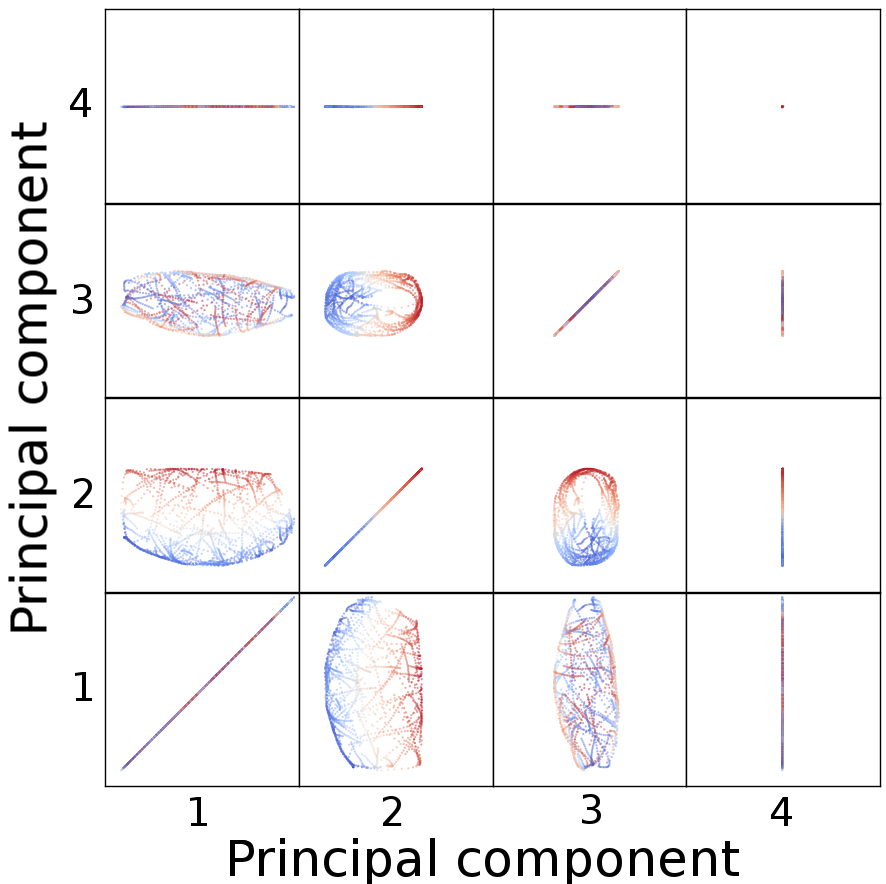}
        \caption{\centering Learned representation\hspace{\textwidth}(moving camera)}
        \label{fig:rep_cartpole_moving}
    \end{subfigure}
    \hfill
    \begin{subfigure}[t]{0.49\columnwidth}
        \includegraphics[width=\columnwidth]{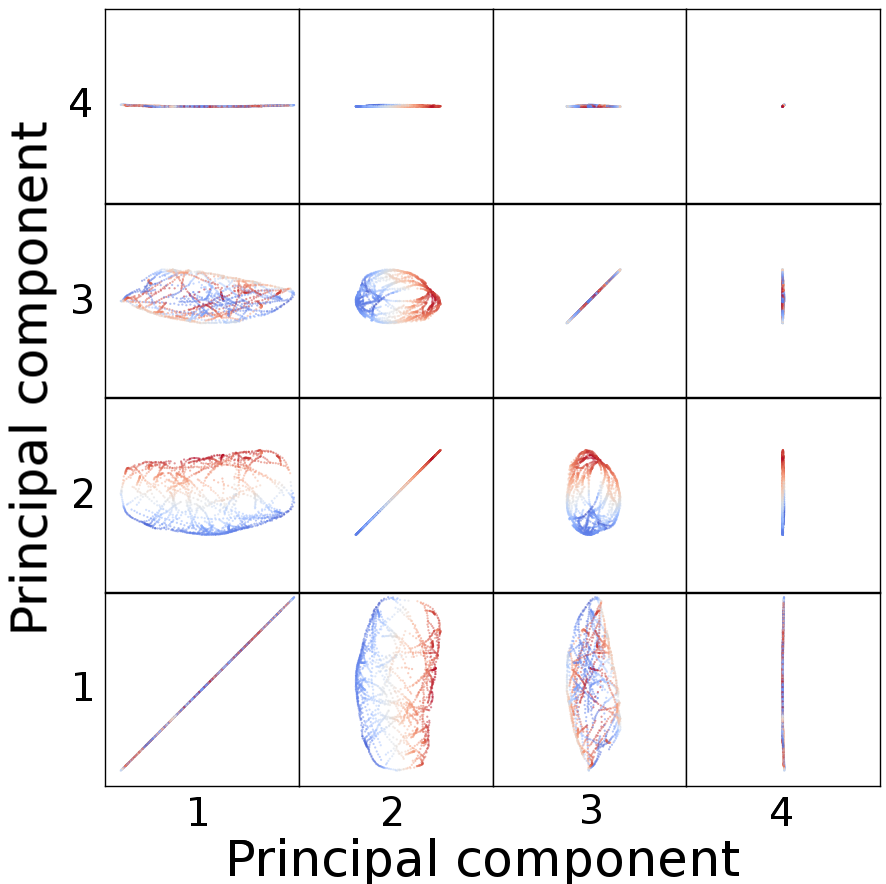}
        \caption{\centering Learned representation\hspace{\textwidth}(static camera)}
        \label{fig:rep_cartpole_static}
    \end{subfigure}
    \\
    \begin{subfigure}[t]{0.49\columnwidth}
        \vspace{0.2cm}
        \includegraphics[width=\columnwidth]{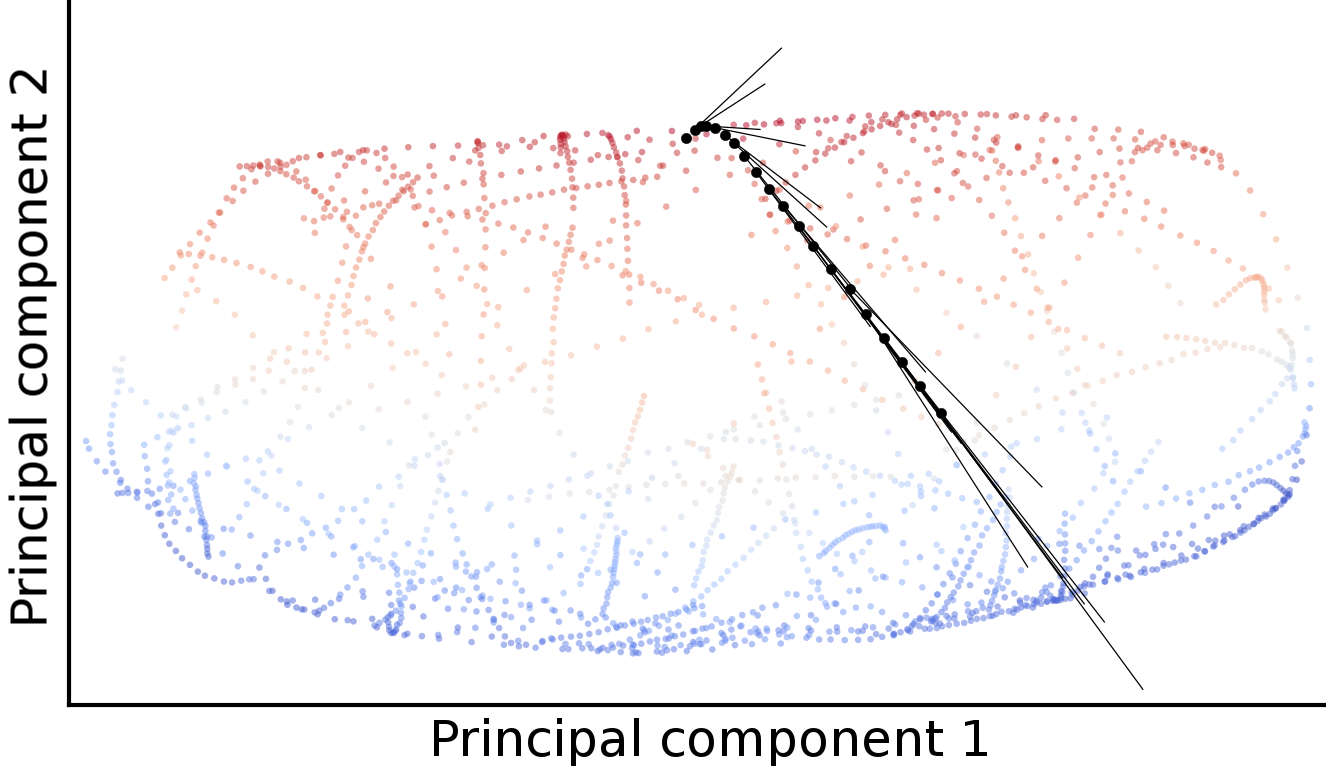}
        \caption{\centering Encoded sequence\hspace{\textwidth}(moving camera)}
        \label{fig:seq_cartpole_moving}
    \end{subfigure}
    \hfill
    \begin{subfigure}[t]{0.49\columnwidth}
        \vspace{0.2cm}
        \includegraphics[width=\columnwidth]{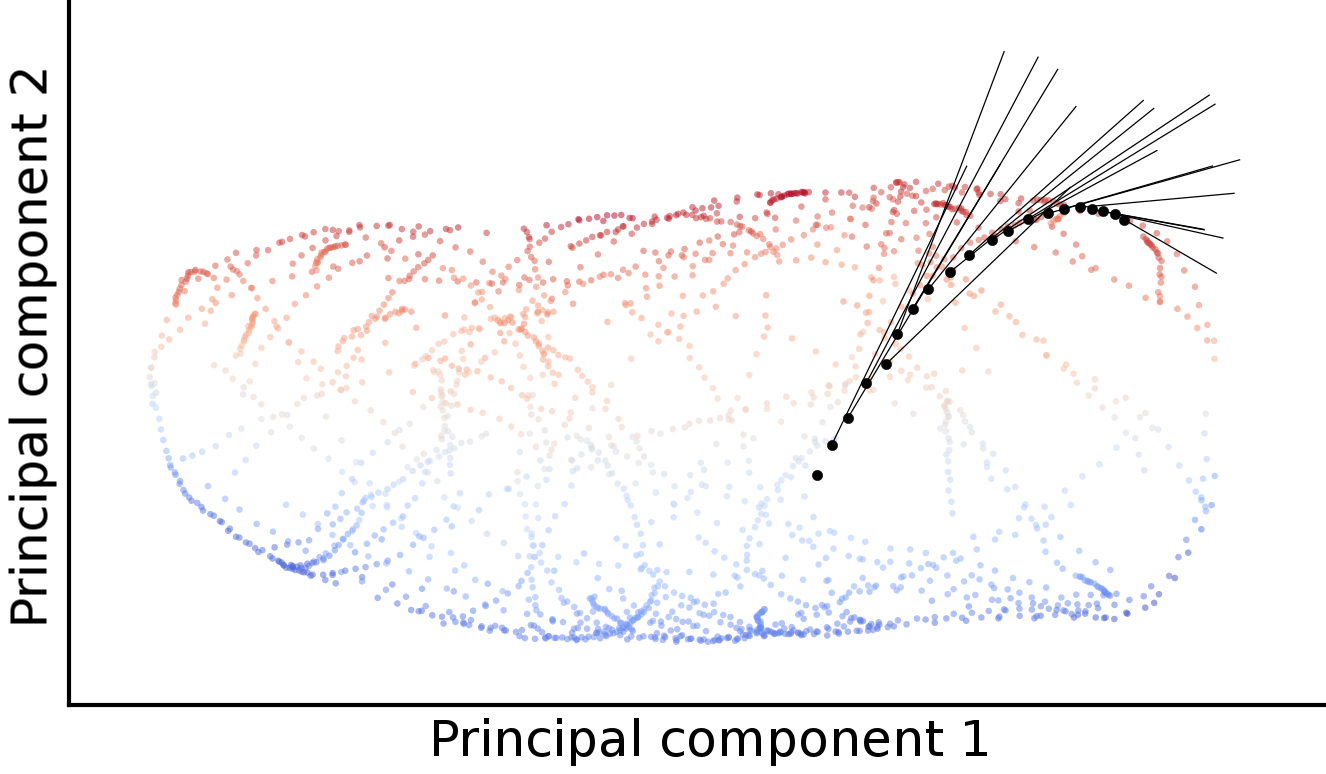}
        \caption{\centering Encoded sequence\hspace{\textwidth}(static camera)}
        \label{fig:seq_cartpole_static}
    \end{subfigure}
    \\
    \begin{subfigure}[t]{0.49\columnwidth}
        \vspace{0.315cm}
        \includegraphics[width=\columnwidth]{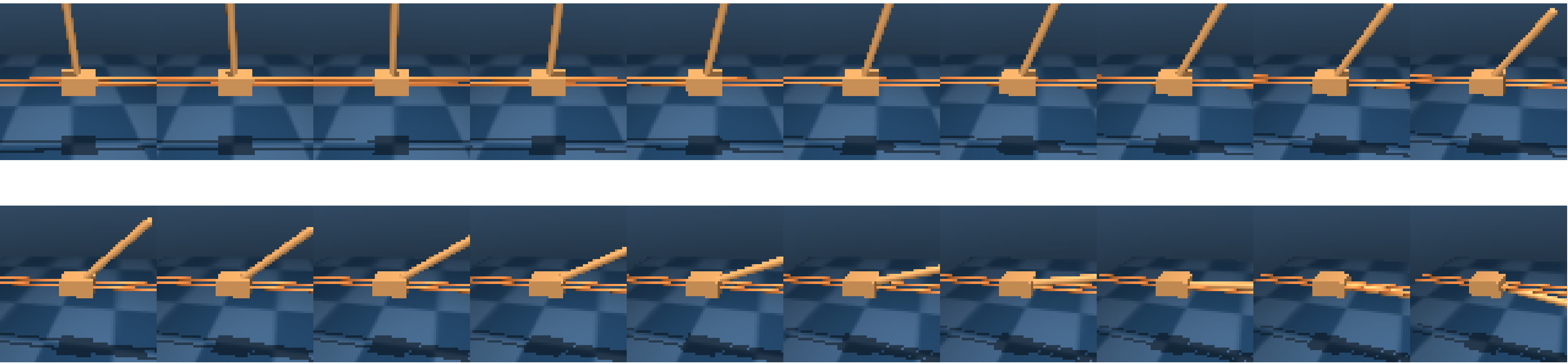}
        \caption{\centering Observation sequence\hspace{\textwidth}(moving camera)}
        \label{fig:obs_cartpole_moving}
    \end{subfigure}
    \hfill
    \begin{subfigure}[t]{0.49\columnwidth}
        \vspace{0.3cm}
        \includegraphics[width=\columnwidth]{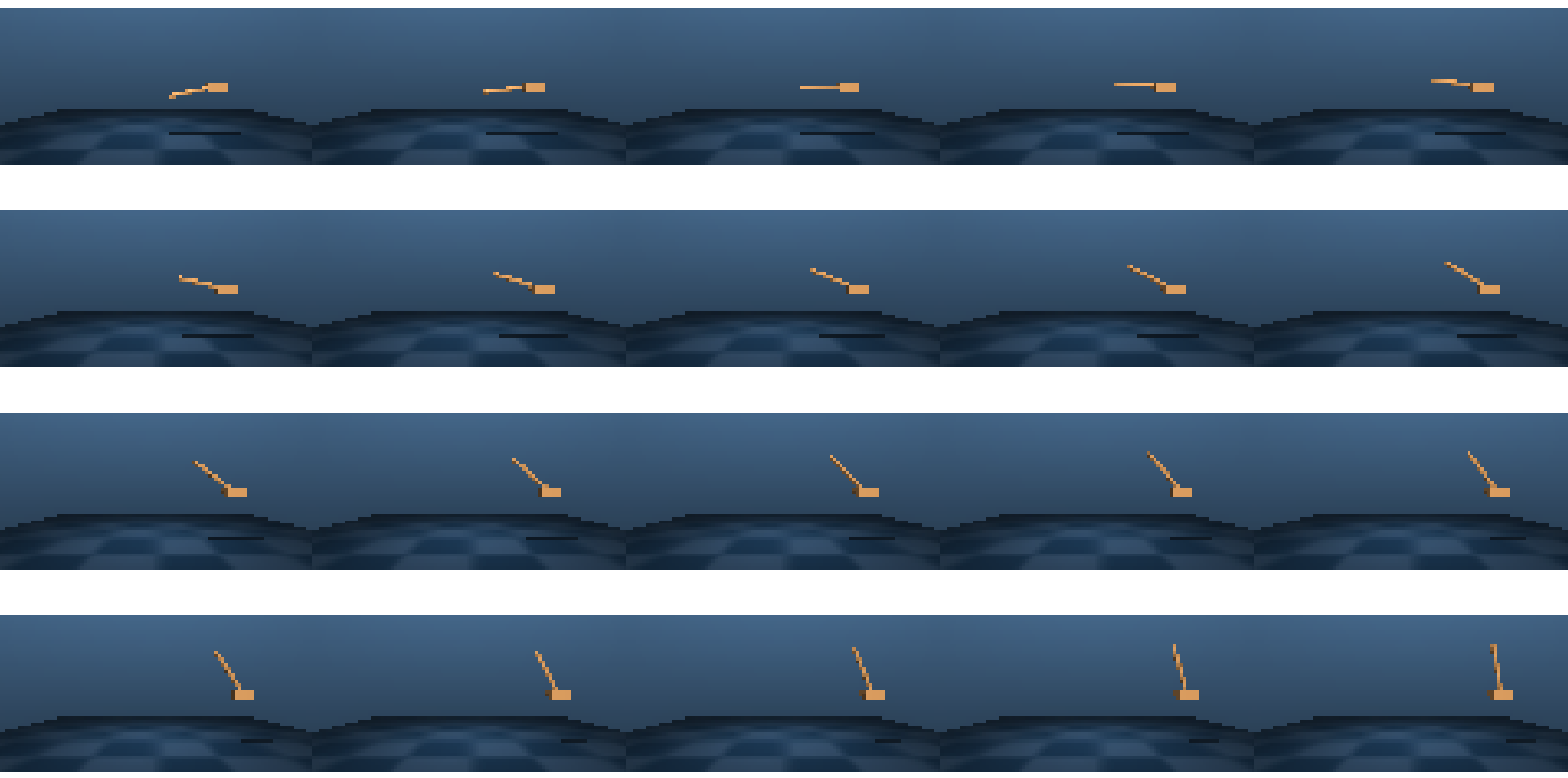}
        \caption{\centering Observation sequence\hspace{\textwidth}(static camera)}
        \label{fig:obs_cartpole_static}
    \end{subfigure}
    \caption{For cart-pole, PVEs learn equivalent state representations from different observations. {\bf Supplementary videos}: learning process for the moving camera \url{http://youtu.be/RKlciWWuJfc} and static camera \url{http://youtu.be/MYxrA1Bw6MU}, learned PVE with the moving camera \url{http://youtu.be/67QZRsLNTAE}.}
    \label{fig:rep_cartpole}
\end{figure}

Here, we compare PVEs on two different observations: 1) using a moving camera that follows the cart by rotating sideways, 2) using a static camera that covers the entire region in which the cart moves. Figure~\ref{fig:rep_cartpole} shows the learned position representations for both perspectives.

This experiment demonstrates how PVEs can learn equivalent internal representations (compare Figs.~\ref{fig:rep_cartpole_moving} and \ref{fig:rep_cartpole_static}) from observations that look very different (Figs.~\ref{fig:obs_cartpole_moving}, \ref{fig:obs_cartpole_static}). For both kinds of observations, the state samples form a tube, the length of which corresponds to the position of the cart, while the circular part represents the position of the pole. Here, the PVE uses three of the five dimensions and thereby discovers the three-dimensional nature of the given task.

The observation sequence from the moving camera (Fig.~\ref{fig:obs_cartpole_moving}) shows the cart moving to the left while the pole falls down on the right side. The PVE represents this trajectory (Fig.~\ref{fig:seq_cartpole_moving}) by moving from the high-reward red region to the blue region, which reflects the movement of the pole, and to the right side, which corresponds to sideways movement of the cart. The observation sequence from the static camera (Fig.~\ref{fig:obs_cartpole_static}) shows the pole swinging up while the cart moves to the right. Accordingly, the encoded trajectory (Fig.~\ref{fig:seq_cartpole_static}) goes to the red region and to the right side (right and left a swapped between these two representations).

\subsubsection{Ball in Cup}

The results for this task are preliminary. The task is challenging due to the movement of the cup, which is inconsistent with some of our robotic priors. The cup is confined to a small region and controlled by the robot allowing rapid movements and changes of direction. The cup can be moved from one end of its position range to the other end in a few time steps. Therefore, the slowness prior does not hold here (unless we sampled observations at a higher frequency). Additionally, the robot can apply large forces on the cup, leading to large accelerations and jerky movements, which are again inconsistent with many of our priors on changes in velocity. As a result, PVEs struggle with encoding the cup, which we will quantify in the following section.

\begin{figure}[t]
    \centering
    \begin{subfigure}[t]{0.48\columnwidth}
        \includegraphics[width=\columnwidth]{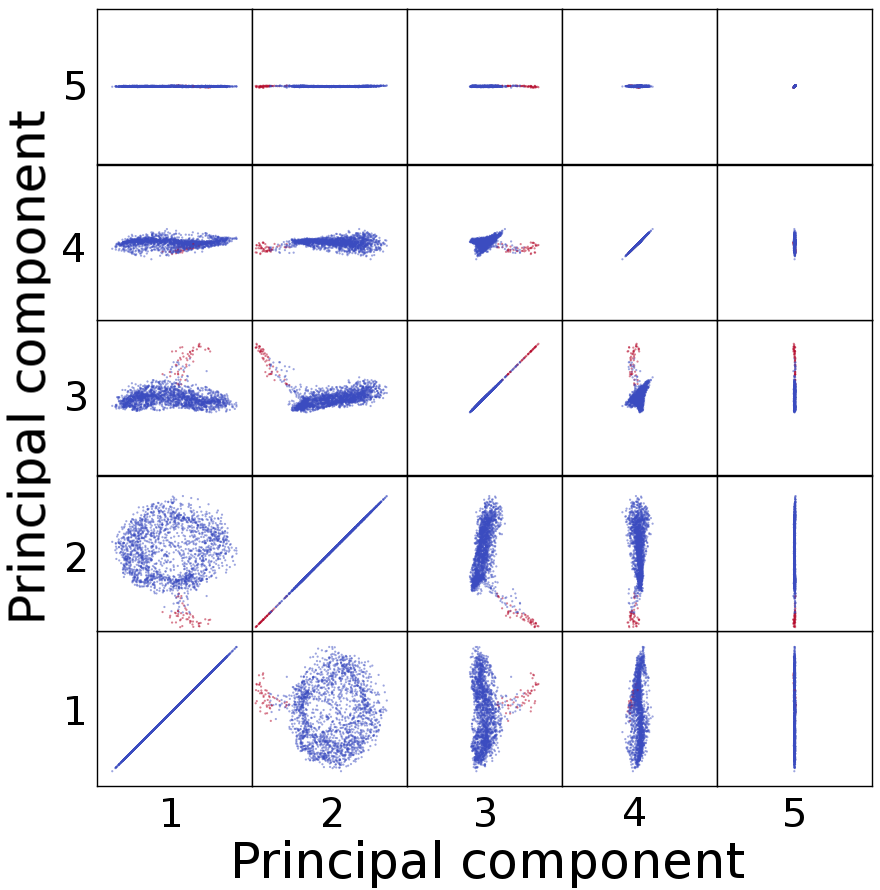}
        \caption{}
    \end{subfigure}
    \hfill
    \begin{subfigure}[t]{0.48\columnwidth}
        \includegraphics[width=\columnwidth]{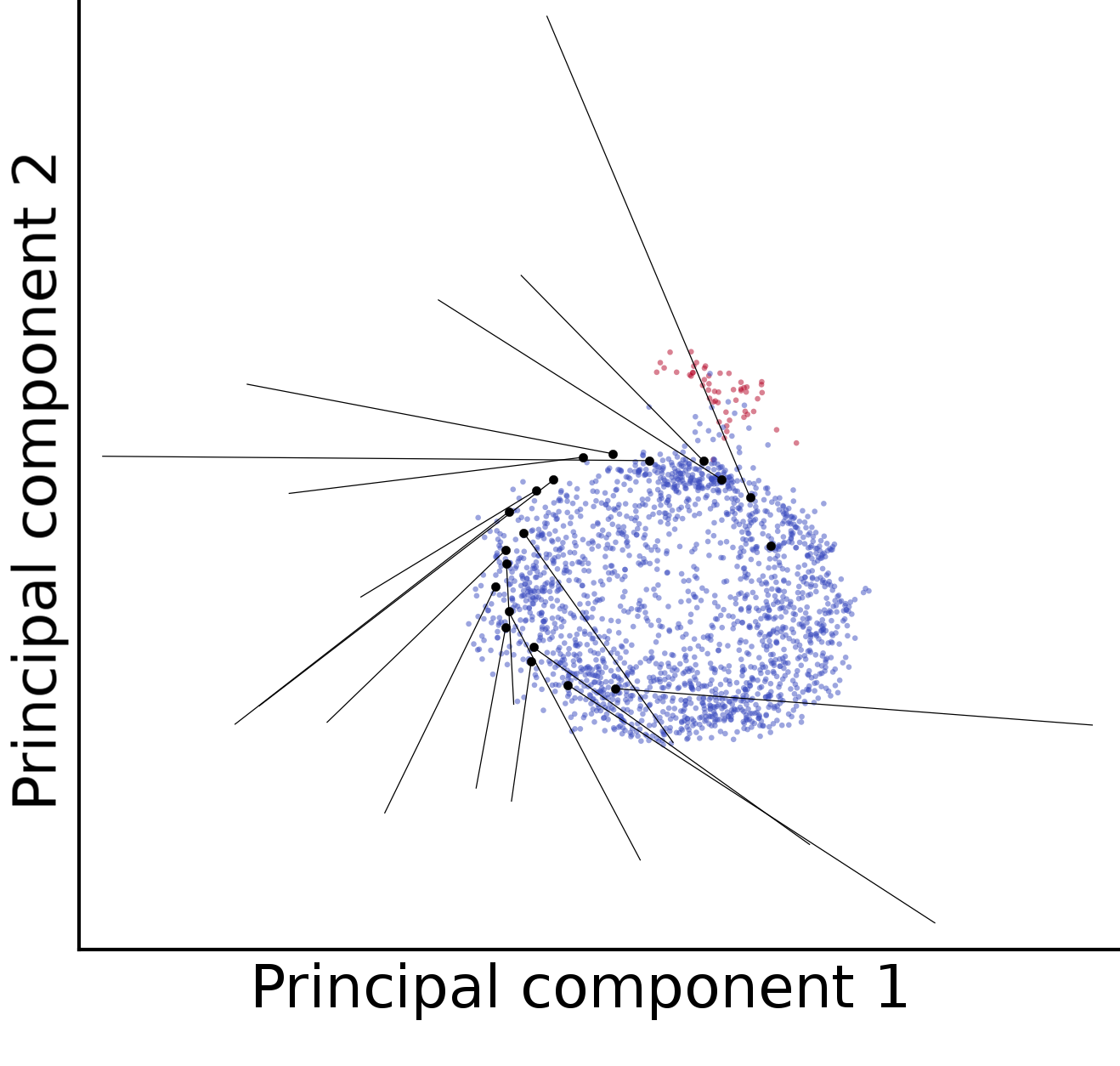}
        \caption{Inverted pendulum}
    \end{subfigure}
    \\
    \begin{subfigure}[t]{\columnwidth}
        \vspace{0.3cm}
        \includegraphics[width=\columnwidth]{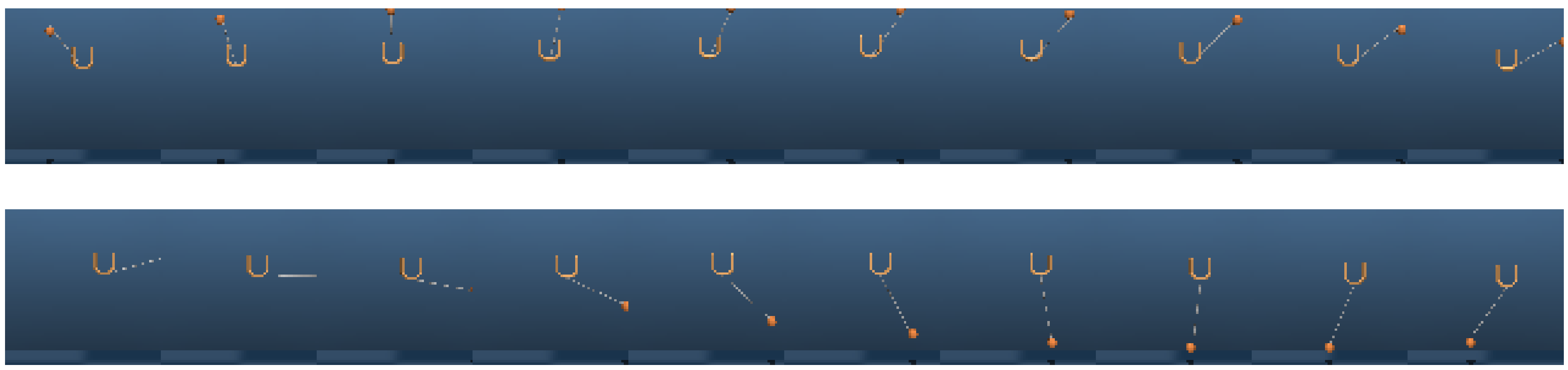}
        \caption{Observation sequence}
        \label{fig:obs_ball_in_cup}
    \end{subfigure}
    \caption{Learned position-velocity representation for ball in cup. {\bf Supplementary videos}: \url{http://youtu.be/3fLaSL8d4TY} shows the learning process, \url{http://youtu.be/lIhEGv5kLFo} demonstrates the learned PVE.}
    \label{fig:ball_in_cup}
\end{figure}

To approach this problem, we added the controllability prior, which enforces that things controlled by the robot are encoded into the state. This improved the resulting state representation (see Fig.~\ref{fig:ball_in_cup}). While the semantics of the state representation are not as clear as for the previous tasks, the representation uses four dimensions, which makes sense for two objects in a plane. Additionally, the goal states (ball in cup) are clearly separated from the other states. As we will see in the following section, the information about the cup is still very noisy, which is probably why reinforcement learning based on PVEs does not reach the same performance as in the other tasks. This result makes the ball in cup task a good candidate for a next step on extending PVEs by revising and adding robotic priors. 

\subsection{Regression to True Positions and Velocities}

To measure the quality of the learned position-velocity representation, we performed regression from the learned postion-velocity state to true positions and velocities of relevant objects. Here, we trained a fully connected neural network with 3 hidden ReLu layers of 256 units each for 200 steps with Adam. We normalized the true positions and velocities and performed supervised learning from the learned position-velocity state to the true features minimizing mean squared error. After training on observations from 1000 times 20 steps, we tested with 100 times 20 test samples. The resulting test errors are shown in Table~\ref{tab:mse}.

\begin{table}[ht]
\centering
\caption{Comparison of mean squared test errors.}
\label{tab:mse}
\begin{subtable}[c]{0.48\columnwidth}
\centering
\begin{tabular}{|l|l|l|}
    \cline{1-2}
    \multicolumn{2}{|c|}{Inverted pendulum} \\
    \cline{1-2}
    $\cos(\theta_{\textrm{pole}})$ & 0.0003 & \multicolumn{1}{c}{} \\
    $\sin(\theta_{\textrm{pole}})$ & 0.0002 & \multicolumn{1}{c}{} \\
    $\dot{\theta}_{\textrm{pole}}$ & 0.0003 & \multicolumn{1}{c}{} \\
    \cline{1-2}
    \multicolumn{2}{c}{} \\
    \hline
    \multicolumn{3}{|c|}{Cart-pole (different cameras)} \\
    \hline
    & moving & static \\
    \hline
    $x_\textrm{cart}$ & 0.0007 & 0.0015  \\
    $\cos(\theta_{\textrm{pole}})$ & 0.0013 & 0.0021 \\
    $\sin(\theta_{\textrm{pole}})$ & 0.0012  & 0.0033 \\
    $\dot{x}_\textrm{cart}$ & 0.0069  & 0.0198 \\
    $\dot{\theta}_{\textrm{pole}}$ & 0.0110  & 0.0264 \\
    \hline
\end{tabular}
\end{subtable}
\hspace{0.25cm}
\begin{subtable}[c]{0.3\columnwidth}
\begin{tabular}{|l|l|l|}
    \hline
    \multicolumn{2}{|c|}{Ball in cup} \\
    \hline
    $x_\textrm{cup}$ & 0.0622 \\
    $y_\textrm{cup}$ & 0.0645 \\
    $x_\textrm{ball}$ & 0.0187 \\
    $y_\textrm{ball}$ & 0.0294 \\
    $\dot{x}_\textrm{cup}$ &  0.6654 \\
    $\dot{y}_\textrm{cup}$ &  0.6372 \\
    $\dot{x}_\textrm{ball}$ & 0.1535 \\
    $\dot{y}_\textrm{ball}$ & 0.2359 \\
    \hline
    \multicolumn{2}{c}{} \\
    \multicolumn{2}{c}{} \\
    \multicolumn{2}{c}{} \\
\end{tabular}
\end{subtable}
\end{table}

When we compare these errors, we find that the errors are lowest for the pendulum task, which makes sense because the range of possible observations is so small in this task, that it is well covered by the training data. For the cart-pole the errors are still very low for position, but higher for the estimated velocities because noise in the position states is increased when computing velocities from finite differences. Also, the errors double when we go from the moving camera setting to the static camera setting. From this difference, we can predict that control should be easier in the first setting. Finally, for ball in cup, the errors are again much larger for the reasons discussed earlier. The estimation of the cup velocity is particularly challenging. 

Note that we performed this regression test to measure how well these properties are encoded in the state. We do not use the state labels for training the representation and we do not use them for learning control. In the following section, we will measure the utility of the learned representation by reinforcement learning performance based on these representations.








\subsection{Enabling Reinforcement Learning}

In this experiment, we learn control for these tasks with neural fitted Q-iteration (NFQ, \cite{NFQ}) based on the encoding learned by PVEs. As a baseline, we use untrained PVEs with randomly initialized encodings in this preliminary work (we will thoroughly compare to other methods in future work). For the policy, we used a fully connected neural network with two hidden layers of 250 sigmoid units per layer. We trained it two times for 30 episodes after each training epoch. We rescaled rewards to be non-positive and ommitted discounting. We repeated actions for multiple time steps (4 for the pendulum and cart-pole tasks, 6 for the ball in cup task).

\begin{figure}[t]
    \begin{subfigure}{0.49\columnwidth}
        \includegraphics[width=\columnwidth]{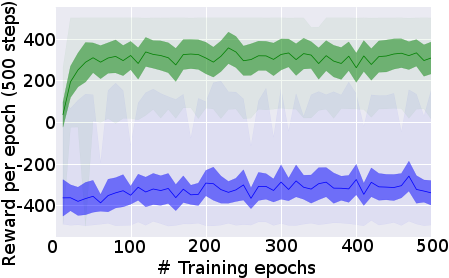}
        \caption{Inverted pendulum}
        \label{fig:lc_pendulum}
    \end{subfigure}
    \hfill
    \begin{subfigure}{0.49\columnwidth}
        \includegraphics[width=\columnwidth]{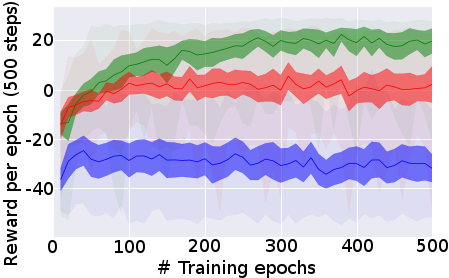}
        \caption{Cart-pole}
        \label{fig:lc_cart_pole}
    \end{subfigure}
    \\
    \begin{subfigure}{0.49\columnwidth}
        \includegraphics[width=\columnwidth]{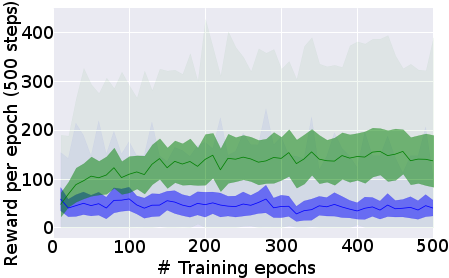}
        \caption{Ball in cup}
        \label{fig:lc_ball_in_cup}
    \end{subfigure}
    \hfill
    \begin{subfigure}{0.49\columnwidth}
        \includegraphics[width=\columnwidth]{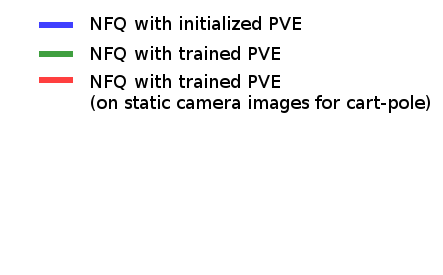}
    \end{subfigure}
    \caption{Reinforcement learning performance for different tasks based on state representations learned by PVEs. Lines show means of 50 trials, darker shading shows standard errors, lighter shading shows range from minimum to maximum values.}
    \label{fig:learning_curves}
\end{figure}

The resulting learning curves are shown in Figure~\ref{fig:learning_curves}. The blue curves show the baselines with random encodings, which do not allow learning any of the three tasks. The green and red curves show reinforcement learning based on PVEs that were trained on a batch of 1000 trajectories of 20 steps. For the inverted pendulum and for the cart-pole task, the green curves reach optimal performance after only 50 and 300 epochs. The red curve, which shows the performance based on the static camera perspective does not reach optimal performance, probably due to the more noisy state estimation discussed in the previous section. At this point, it is not clear whether this issue comes from the low resolution in the input or from the fact that the position of the pole and the cart are more strongly coupled in these observations which makes learning the state encoding more difficult. Lastly, for the ball in cup task, the learned control beats the baseline consistently and (as the light green maximum shading shows) more successful control using the learned representation is possible. But due to the noisy state estimation, this is not sufficient for solving the task consistently. Future work could start from here and investigate which priors are missing to solve this and more realistic robotic tasks.

\section{Conclusion and Future Work}
\label{sec:conclusion}
We have presented position-velocity encoders (PVEs), which are able to learn state representations structured into a position and a velocity part without supervision and without requiring image reconstruction. The keys to PVEs are to constrain the model to estimate velocities in the correct way from positions and to train the position encoder by optimizing consistency with robotic priors, which are specific to positions, velocities, and accelerations. We have shown how structuring the state space into positions and velocities opens up new opportunities for formulating useful constraints and learning objectives. In future research, we will work towards adding further structure into the state space, revising and extending the list of robotic priors, and combining these approaches with end-to-end reinforcement learning.

\balance
\bibliographystyle{plain}
\bibliography{references}
\end{document}